\documentclass{article}

\PassOptionsToPackage{numbers, compress}{natbib}

\usepackage[preprint]{neurips_2019}

\usepackage[utf8]{inputenc} 
\usepackage[T1]{fontenc}    
\usepackage{hyperref}       
\usepackage{url}            
\usepackage{booktabs}       
\usepackage{amsfonts}       
\usepackage{nicefrac}       
\usepackage{microtype}      
\usepackage{graphicx}
\usepackage{subfigure}
\usepackage{wrapfig}
\usepackage{amsmath}
\usepackage{amssymb}
\usepackage{amsthm}
\usepackage{bbm}
\usepackage{dsfont}
\usepackage{makecell}
\usepackage{paralist}
\usepackage{algorithm}
\usepackage{algorithmic}
\usepackage{multirow}

\def\b{{\bf b}}

\def\d{{\bf d}}

\def\f{{\bf f}}

\def\g{{\bf g}}

\def\h{{\bf h}}

\def\X{{\bf X}}

\def\x{{\bf x}}
\def\y{{\bf y}}

\def\q{{\bf q}}

\def\W{{\bf W}}
\def\w{{\bf w}}
\def\X{{\bf X}}

\def\0{{\bf 0}}
\def\1{{\bf 1}}

\def\RB{{\mathbb R}}

\def\LM{{\mathcal L}}

\def\sgn{\text{sign}}

\def\st{\text{s.t.}}

\newtheorem{proposition}{Proposition}

\title{Deep Multi-Index Hashing for Person Re-Identification}

\author{%
  Ming-Wei Li, Qing-Yuan Jiang and Wu-Jun Li\\
  National Key Laboratory for Novel Software Technology\\
  Collaborative Innovation Center of Novel Software Technology and Industrialization\\
  Department of Computer Science and Technology, Nanjing University, China\\
  \texttt{\{jiangqy,limw\}@lamda.nju.edu.cn, liwujun@nju.edu.cn} \\
}

\begin{document}

\maketitle
\begin{abstract}
Traditional person re-identification~(ReID) methods typically represent person images as real-valued features, which makes ReID inefficient when the gallery set is extremely large. Recently, some hashing methods have been proposed to make ReID more efficient. However, these hashing methods will deteriorate the accuracy in general, and the efficiency of them is still not high enough. In this paper, we propose a novel hashing method, called \underline{d}eep \underline{m}ulti-\underline{i}ndex \underline{h}ashing~(\mbox{DMIH}), to improve both efficiency and accuracy for ReID. DMIH seamlessly integrates \mbox{multi-index} hashing and \mbox{multi-branch} based networks into the same framework. Furthermore, a novel \mbox{block-wise} \mbox{multi-index} hashing table construction approach and a \mbox{search-aware} \mbox{multi-index}~(SAMI) loss are proposed in \mbox{DMIH} to improve the search efficiency. Experiments on three widely used datasets show that DMIH can outperform other \mbox{state-of-the-art} baselines, including both hashing methods and \mbox{real-valued} methods, in terms of both efficiency and accuracy.
\end{abstract}

\section{Introduction}
Person re-identification~(ReID)~\cite{DBLP:conf/eccv/SunZYTW18, DBLP:conf/cvpr/ChangHX18, DBLP:conf/nips/GeLZYYWL18} has attracted much attention in computer vision. For a given probe person, the goal of ReID is to retrieve~(search) in the gallery set for pedestrian images containing the same individual in a \mbox{cross-camera} mode. Recently, \mbox{ReID} has been widely used in many real applications including \mbox{content-based} video retrieval, video surveillance, and so on.

Existing ReID methods can be divided into two main categories~\cite{DBLP:journals/corr/ZhengYH16}. One category focuses on utilizing \mbox{hand-crafted} features to represent person images, especially for most early ReID approaches~\cite{DBLP:conf/cvpr/KostingerHWRB12,DBLP:conf/cvpr/ZhaoOW14,DBLP:conf/cvpr/LiaoHZL15}. The other category~\cite{DBLP:conf/mm/WangYCLZ18, DBLP:conf/eccv/SunZYTW18, DBLP:conf/cvpr/ChangHX18} adopts deep learning architectures to extract features. Most of these existing methods, including both deep methods and non-deep methods, typically represent person images as real-valued features. This real-valued feature representation makes \mbox{ReID} inefficient when the gallery set is extremely large, due to high computation and storage cost during the retrieval~(search) procedure.

Recently, hashing~\cite{DBLP:conf/nips/WeissTF08, DBLP:conf/nips/LiSHT17, DBLP:conf/nips/SuZHT18, DBLP:conf/nips/LiuMKC14, DBLP:conf/icml/LiLSHD13, DBLP:conf/icml/YuKGC14, DBLP:conf/icml/LiuWKC11, DBLP:conf/icml/DaiGKHS17, DBLP:conf/icml/NorouziF11, DBLP:conf/icml/WangKC10} has been introduced into ReID community for efficiency improvement due to its low storage cost and fast query speed. The goal of hashing is to embed data points into a Hamming space of binary codes where the similarity in the original space is preserved. Several hashing methods have been proposed for ReID~\cite{DBLP:journals/tip/ZhangLZZZ15,DBLP:conf/ijcai/ZhengS16,DBLP:conf/cvpr/ChenWQLS17,DBLP:journals/tip/ZhuKZFT17}. In~\cite{DBLP:conf/ijcai/ZhengS16}, deep regularized similarity comparison hashing~(DRSCH) was designed by combining \mbox{triplet-based} formulation and \mbox{bit-scalable} binary codes generation. In~\cite{DBLP:conf/ijcai/ZhengS16}, \mbox{cross-view} binary identifies~(CBI) was learned by constructing two sets of discriminative hash functions. In~\cite{DBLP:conf/cvpr/ChenWQLS17}, \mbox{cross-camera} semantic binary transformation~(CSBT) employed subspace projection to mitigate \mbox{cross-camera} variations. In~\cite{DBLP:journals/tip/ZhuKZFT17}, \mbox{part-based} deep hashing~(PDH) was proposed to incorporate \mbox{triplet-based} formulation and image partitions to learn \mbox{part-based} binary codes.  Among these methods, CBI and CSBT focus on designing models to learn binary codes by using \mbox{hand-crafted} features. DRSCH and PDH are deep hashing methods which try to integrate deep feature learning and hash code learning into an \mbox{end-to-end} framework. Recent efforts~\cite{DBLP:conf/nips/LiSHT17, DBLP:conf/nips/SuZHT18, DBLP:journals/tip/ZhuKZFT17} show that the deep hashing methods can achieve better performance than hand-crafted feature based hashing methods.

However, existing \mbox{hashing} methods usually cannot achieve satisfactory performance for ReID. Exhaustive linear search based on Hamming ranking cannot handle \mbox{large-scale} dataset. More specifically, although one can adopt hash lookup to achieve \mbox{sub-linear} query speed, they~\cite{DBLP:journals/tip/ZhuKZFT17, DBLP:conf/cvpr/ChenWQLS17} usually need long binary codes to achieve reasonable accuracy due to the high complexity in ReID. In this situation, the retrieval speed will become extremely slow because the number of hash bins that need to be retrieved increases exponentially as code length increases. Hence, although existing hashing methods can achieve faster speed than traditional real-valued ReID methods, these hashing methods will typically deteriorate the accuracy because the binary code cannot be too long. Furthermore, the efficiency of existing hashing methods is still not high enough.

In this paper, we propose a novel hashing method, called \underline{d}eep \underline{m}ulti-\underline{i}ndex \underline{h}ashing~(DMIH), to improve both retrieval efficiency and accuracy for ReID. Our main contributions are summarized as follows.
\begin{inparaenum}[1)]
\item DMIH seamlessly integrates multi-index hashing~(\mbox{MIH})~\cite{DBLP:journals/pami/0002PF14} and multi-branch based networks into the same framework. In DMIH, feature learning procedure and hash code learning procedure can facilitate each other. To the best of our knowledge, DMIH is the first hashing based ReID method to integrate \mbox{multi-index} hashing and deep feature learning into the same framework.
\item In DMIH, a novel block-wise multi-index hashing table construction approach and a search-aware multi-index~(SAMI) loss are proposed to improve the retrieval efficiency.
\item Experiments on three widely used datasets show that DMIH can outperform other state-of-the-art baselines, including both hashing methods and real-valued methods, in terms of both efficiency and accuracy.
\end{inparaenum}

\section{Related Work}\label{sec:related-work}

\paragraph{Multi-Index Hashing} In real applications, when facing long binary codes, hash lookup will suffer from low retrieval speed due to large number of hash bins that need to be retrieved. \mbox{Multi-index} hashing~(MIH)~\cite{DBLP:journals/pami/0002PF14}, which can enable efficient $k$-nearest neighbors search\footnote{Please note that here the $k$-nearest neighbors are defined based on Hamming distance~\cite{DBLP:journals/pami/0002PF14}.} for long codes is proposed to deal with this situation. MIH divides the long binary codes into several disjoint but consecutive \mbox{sub-binary} codes and builds multiple hash tables on shorter code substrings, which can enormously reduce the number of hash bins to be retrieved and improve the search efficiency.

However, MIH is based on the assumption that the binary codes should be distributed balanced between \mbox{sub-binary} codes, which is usually not satisfied in real applications~\cite{DBLP:conf/mm/ZhangGZL11}. So the time performance of MIH will be adversely affected when dealing with unbalanced distributed codes. Our method learns to adjust the distribution of binary codes by minimizing SAMI loss to enhance the time performance of MIH.

\paragraph{Multi-Branch Architectures}  Multi-branch based networks~\cite{DBLP:conf/cvpr/SzegedyLJSRAEVR15, DBLP:conf/cvpr/HeZRS16} have been widely exploited in computer vision tasks. Recently, ``grouped convolution''~\cite{DBLP:conf/cvpr/XieGDTH17, DBLP:conf/cvpr/HuangLMW18} has been proposed to construct multi-branch architectures. These building blocks can achieve stronger modeling capacity. In ReID, due to the cross-camera variations, the partial information is significant to improve the discriminative performances. Multi-branch based networks have been used to learn discriminative information with various granularities in previous works~\cite{DBLP:journals/tip/ZhuKZFT17, DBLP:conf/mm/WangYCLZ18, DBLP:conf/eccv/SunZYTW18, DBLP:conf/cvpr/ChangHX18}.

\section{Notation and Problem Definition}\label{sec:notation}

\subsection{Notation}
We use boldface lowercase letters like $\w$ to denote vectors and boldface uppercase letters like $\W$ to denote matrices. $\Vert\w\Vert_2$ denotes the $L_2$-norm for the vector $\w$. $[\cdot]_{+}$ is defined as $[x]_+=\max\{0,x\}$. For an integer $C$, we use $[C]$ to denote the set $\{1,2,\dots,C\}$. $\sgn(\cdot)$ is an element-wise sign function where $\sgn(x) = 1$ if $x\geq 0$ else $\sgn(x)=-1$. 
Furthermore, $\Vert\b_i-\b_j\Vert_H$ denotes the Hamming distance between two binary vectors $\b_i$ and $\b_j$, i.e., $\Vert\b_i-\b_j\Vert_H=(R-\b_i^T\b_j)/2$. Here, $R$ is the code length of $\b_i$ and $\b_j$.

\subsection{Hashing based ReID}
Assume that we have $n$ training samples which are denoted as $\X=\{\x_i\}_{i=1}^n$. Furthermore, person identities for images are also available and denoted as $\y=\{y_i\;\vert\;y_i\in[C]\}_{i=1}^n$, where $C$ denotes the number of persons in the training set. Our target is to learn a deep hash function $H(\x)\in \{-1, +1\}^R$, which can transform the person images to binary codes with $R$ bits.


\section{Deep Multi-Index Hashing for ReID}\label{sec:model}

\subsection{Model}
\begin{figure}[!htb]
\begin{center}
\includegraphics[scale=0.275]{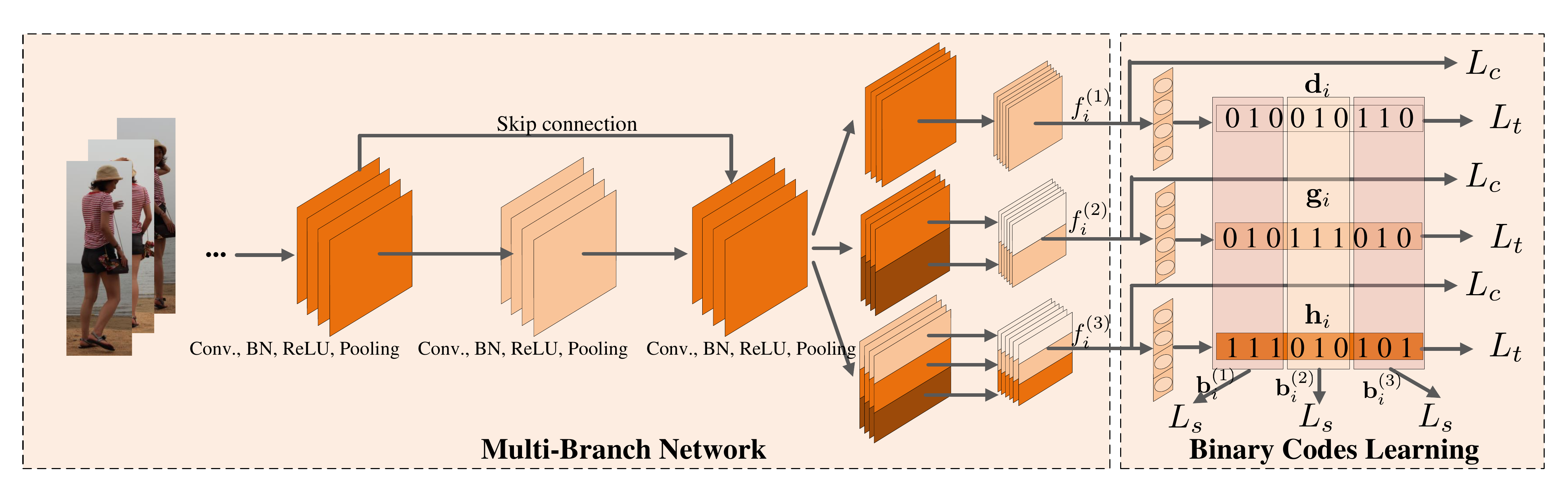}\vspace{-10pt}
\caption{The deep multi-index hashing~(DMIH) framework for ReID.}\vspace{-15pt}
\label{fig:framework}
\end{center}
\end{figure}

The DMIH model is illustrated in Figure~\ref{fig:framework}, which is an \mbox{end-to-end} deep learning framework containing two components, i.e., \mbox{multi-branch} network part and binary codes learning part. Furthermore, a novel \mbox{block-wise} \mbox{multi-index} hashing table construction approach and a novel \mbox{search-aware} \mbox{multi-index}~(SAMI) loss are developed to improve search efficiency.

\paragraph{Multi-Branch Network Part} As a variety of deep methods for ReID~\cite{DBLP:journals/tip/ZhuKZFT17, DBLP:conf/mm/WangYCLZ18, DBLP:conf/eccv/SunZYTW18, DBLP:conf/cvpr/ChangHX18} have demonstrated that multi-branch architectures based network can learn more discriminative features, we adopt the \mbox{multi-granularity} network~(MGN) architecture~\cite{DBLP:conf/mm/WangYCLZ18} as the feature learning part of DMIH. This architecture integrates global and local features to get more powerful pedestrian representations. The MGN architecture is shown in the left part of Figure~\ref{fig:framework}, which contains a \mbox{ResNet-50}~\cite{DBLP:conf/cvpr/HeZRS16} network and three branches. The upper branch without any partition information learns the global feature representations. The middle and lower branches uniformly split feature maps into several stripes in horizontal orientation to learn the local feature representations. For a given pedestrian image $\x_i$, the output of all the three branches are denoted as $\{\f_i^{(1)}, \f_i^{(2)}, \f_i^{(3)}\}$, which contains the global and local feature representations. Please note that DMIH is general enough to adopt other multi-branch architectures since our objective is to improve the retrieval efficiency and accuracy, rather than designing a new multi-branch building block. In other words, our method is an extensive learning algorithm, which is independent of specific network architectures.

\paragraph{Binary Codes Learning Part} The principle of binary codes learning is to preserve the similarity of samples. We use \mbox{triplet-based} loss to achieve this goal, which has been proved to be effective in deep ReID tasks~\cite{DBLP:journals/tip/ZhangLZZZ15,DBLP:journals/tip/ZhuKZFT17,DBLP:conf/mm/WangYCLZ18}. Specifically, for the $i$-th input $\x_i$, we add a \mbox{fully-connected} layer after each branch of network as a hash layer to project the global and local features, i.e., $\{\f_i^{(1)}, \f_i^{(2)}, \f_i^{(3)}\}$, into $\RB^r$. Then we employ the function $\sgn(\cdot)$ to get its corresponding binary codes $\{\d_i, \g_i, \h_i\}$, where $\d_i,\g_i,\h_i \in\{-1,+1\}^r$ and $r$ denotes the code length of each \mbox{sub-binary} code.

Then a triplet loss function is imposed on $\{\d_i, \g_i, \h_i\}$. For example, the loss function for a mini-batch $\{\d_i\}^N_{i=1}$ with $N$ samples can be formulated as follows:
\begin{align}
L_{t}(\{\d_i\}_{i=1}^N) = \frac{1}{N}\sum_{i=1}^N\Big[& \alpha+\max\Vert\d_{i}-\d^{+}_{i}\Vert_H-\min\Vert\d_{i}-\d_{i}^{-}\Vert_H\Big]_{+},\nonumber
\end{align}
where $\d_i, \d_i^{+}, \d_i^{-}$ respectively represent the generated binary codes from anchor, positive and negative samples, $\alpha$ is the margin \mbox{hyper-parameter}. Here the pedestrian who has the same/different identity with the anchor is the positive/negative sample. For each pedestrian image in a \mbox{mini-batch}, we treat it as an anchor and build the corresponding triplet input by choosing the furthest positive and the closest negative samples in the same batch. This improved version of the \mbox{batch-hard} triplet loss enhances the robustness in metric learning ~\cite{DBLP:journals/corr/HermansBL17}, and improves the accuracy at the same time.
Then we can get the following total triplet loss function:
\begin{align}
\label{loss:totaltriplet}
\LM_{t}(\{\d_i,\g_i,\h_i\}_{i=1}^N)=&L_{t}(\{\d_i\}_{i=1}^N)+L_{t}(\{\g_i\}_{i=1}^N)+L_{t}(\{\h_i\}_{i=1}^N).
\end{align}

In order to learn more discriminative binary codes, we explore cross entropy loss for classification on the outputs of each branch. Specifically, we utilize the formula: $L_{c}(\{\f_i^{(j)}\}_{i=1}^N) = -\frac{1}{N}\sum_{i=1}^N{\text{softmax}}(\f_i^{(j)})$.

Then we can get the total classification loss function:
\begin{align}
\label{loss:totalsoftmax}
\LM_{c}(\{\f_i^{(1)},\f_i^{(2)},\f_i^{(3)}\}_{i=1}^N)=&\sum_{j=1}^3L_{c}(\{\f_i^{(j)}\}_{i=1}^N).
\end{align}

\paragraph{Multi-Index Hashing Tables Construction} MIH supposes that each binary code $\b$ with $R$ bits is partitioned into $m$ disjoint isometric \mbox{sub-binary} codes $\b^{(1)},\cdots,\b^{(m)}$. Given a query code $\q$, we aim to find all binary codes with the Hamming distance from $\q$ being $k$. We call them $k$-neighbors. Let $k'=\lfloor k/m\rfloor$ and $a=k-mk'$. According to the Proposition~\ref{pro:mih} proved in~\cite{DBLP:journals/pami/0002PF14}, we only need to search the first $a+1$ hash tables at the radius of $k'$ and the remaining $m-(a+1)$ hash tables at the radius of $k'-1$ to construct a candidate set when performing retrieval procedure for a given query. After that, we remove the points which are not $k$-neighbors from the candidate set by measuring full Hamming distance.
\begin{proposition}
\label{pro:mih}
if $\Vert\b-\q\Vert_H \leq k = mk' + a$, then $\exists~1 \leq z \leq a + 1~\st\;\Vert\b^{(z)}-\q^{(z)}\Vert_H \leq k'$ or $\exists~a + 1 < z \leq m~\st\;\Vert\b^{(z)}-\q^{(z)}\Vert_H \leq k' - 1.$
\end{proposition}

\begin{wrapfigure}{r}{0.5\textwidth}\vspace{-15pt}
\centering
\includegraphics[width=0.49\textwidth]{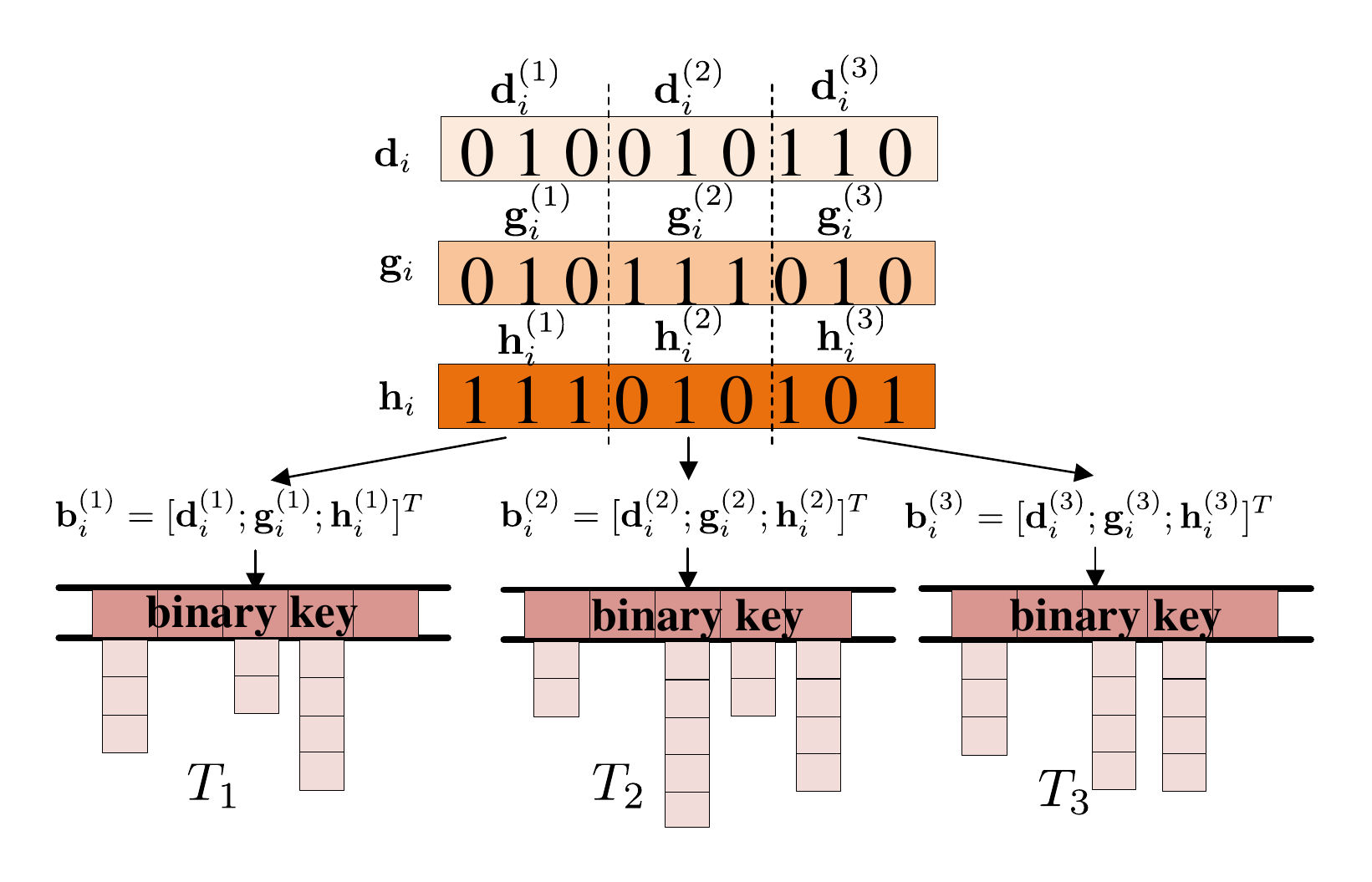}\vspace{-10pt}
\caption{Block-wise MIH tables construction. For each binary code $\{\d_i, \g_i, \h_i\}$, we partition it evenly into $m$ disjoint \mbox{sub-binary} codes and concatenate the corresponding parts to construct $m$ hash tables. Here, we set $m=3$ as an example.}\vspace{-15pt}
\label{fig:MIH}
\end{wrapfigure}

Once we get the learned binary codes, one way to construct MIH tables is to divide the total binary codes into $m$ hash tables, where the total binary codes is defined as $[\d_i;\g_i;\h_i]^T$. By doing so, each hash table might suffer from \mbox{single-granularity} binary codes problem, and thus leads to large difference between different hash tables. To alleviate this situation, we design a novel \mbox{block-wise} MIH tables construction strategy, which is shown in Figure~\ref{fig:MIH}. Specifically, we divide the learned binary codes $\{\d_i,\g_i,\h_i\}$ into $m$ disjoint \mbox{sub-binary} codes \emph{separately}. Then we concatenate all the $j$-th \mbox{sub-binary} codes to construct $\b_i^{(j)}$ for the $j$-th hash table, i.e., $\b_i^{(j)}=[\d_i^{(j)};\g_i^{(j)};\h_i^{(j)}]^T$. That is to say, this \mbox{block-wise} partition strategy can ensure that each hash table contains \mbox{multi-granularity} binary codes.
\paragraph{Search-Aware Multi-Index Loss} Based on the retrieval procedure of MIH tables, we propose a novel loss, called \mbox{search-aware} \mbox{multi-index}~(SAMI) loss, to give feedback to the training procedure. Firstly, we define the binary codes for $\x_i$ and $\x_j$ as $\b_i$ and $\b_j$, respectively. And we use $\b_i^{(l)}$ and $\b_j^{(l)}$ to denote the corresponding \mbox{sub-binary} codes in the $l$-th hash tables. Then we define the Hamming distance between $\b_i^{(l)}$ and $\b_j^{(l)}$ as $\Theta_{ij}^{(l)} = \Vert\b_i^{(l)}-\b_j^{(l)}\Vert_H$. As the first $a+1$ hash tables will be searched firstly, we hope $\Theta_{ij}^{(l)}$ is larger than $\Theta_{ij}^{(l+1)}$ as much as possible. Then we can get the following SAMI loss function:
\begin{small}
\begin{align}
\label{loss:hl-abc}
\LM_{s}(\{\b_i^{(1)},\dots,\b_i^{(m)}\}_{i=1}^N) = \frac{1}{N^2(m-1)}\sum_{i,j=1}^{N}\sum_{l=1}^{m-1}\Big[\Theta_{ij}^{(l+1)} -  \Theta_{ij}^{(l)} \Big]_{+}.
\end{align}
\end{small}

By minimizing the above loss function, we can avoid the situation where the false data points are chosen into the candidate set too early when we utilize MIH tables to perform the retrieval procedure. Here the false data points are those data points whose Hamming distance to a given query point $\q$ is larger than the distance we need to retrieval, but the Hamming distance between its \mbox{sub-binary} codes and the sub-binary codes of the query in the first-searched hash tables is small. Then we can reduce the number of points which are actually not $k$-neighbors but are added into the candidate set. As a result, the time for measuring full Hamming distance and removing the points which are not $k$-neighbors will be saved.

Then we can get the final objective function for DMIH by combining~(\ref{loss:totaltriplet}), (\ref{loss:totalsoftmax}) and (\ref{loss:hl-abc}), which is formulated as follows:
\begin{align}
\min\;\LM=\LM_{t}(\{\d_i,\g_i&,\h_i\}_{i=1}^N)+\beta\LM_{c}(\{\f_i^{(1)},\f_i^{(2)},\f_i^{(3)}\}_{i=1}^N)+\gamma\LM_{s}(\{\b_i^{(1)},\dots,\b_i^{(m)}\}_{i=1}^N)\nonumber\\
&\hspace{2pt}\st\;\d_i,\g_i,\h_i\in\{-1,+1\}^r,\forall i\in\{1,\dots, N\},\label{obj:DMIH}
\end{align}
where $\beta, \gamma$ are hyper-parameters.

\subsection{Learning}
The objective function in~(\ref{obj:DMIH}) is NP-hard due to the binary constraint. One common approach to avoid this problem is to use relaxation strategy~\cite{DBLP:conf/ijcai/LiWK16,DBLP:conf/iccv/CaoLWY17}. In this paper, we also adopt this strategy to avoid this NP-hard problem. Specifically, we utilize $\tanh(\cdot)$ to approximate the $\sgn(\cdot)$ function.

Then we can reformulate the problem in~(\ref{obj:DMIH}) as follows:
\begin{align}
\min\;\LM=\LM_{t}(\{\widetilde\d_i,\widetilde\g_i&,\widetilde\h_i\}_{i=1}^N)+\beta\LM_{c}(\{\f_i^{(1)},\f_i^{(2)},\f_i^{(3)}\}_{i=1}^N)+\gamma\LM_{s}(\{\widetilde\b_i^{(1)},\dots,\widetilde\b_i^{(m)}\}_{i=1}^N)\nonumber\\
&\hspace{2pt}\st\;\widetilde\d_i,\widetilde\g_i,\widetilde\h_i\in[-1,+1]^r,\forall i\in\{1,\dots, N\},\label{obj:relaxDMIH}
\end{align}
where $\widetilde\d_i,\widetilde\g_i,\widetilde\h_i$ denote the continuous codes after relaxation.

Now we can use back propagation to learn the parameters in~(\ref{obj:relaxDMIH}). The learning algorithm for our DMIH is summarized in Algorithm~\ref{alg:DMIH}.
\begin{algorithm}[tb]
\caption{Learning algorithm for DMIH}
\label{alg:DMIH}
\begin{algorithmic}
\REQUIRE
    $\X=\{\x_i\}_{i=1}^n$: images for training; $\y$: person identities for training images; $R$: code length.
\ENSURE
    Parameters of deep neural networks.
\STATE {\bf Procedure}: Initialize parameters of DNN, maximum iteration number $T$, mini-batch size $N$.
\FOR {$t=1\to T$}
    \FOR {$epoch=1\to \lceil n/N \rceil$}
        \STATE Randomly sample $N$ samples from $\X$ to construct a mini-batch $\{\x_i\}^N_{i=1}$.
        \STATE $\forall \x_i\in\{\x_i\}^N_{i=1}$, calculate $\{\f_{i}^{(1)},\f_{i}^{(2)},\f_{i}^{(3)}\}$ and $\{\widetilde\d_i,\widetilde\g_i,\widetilde\h_i\}$ by forward propagation.
        \STATE For mini-batch $\{\x_i\}^N_{i=1}$, calculate corresponding gradient according to loss function in~(\ref{obj:relaxDMIH}).
        \STATE Update the parameters of deep neural network based on the gradient.
    \ENDFOR
\ENDFOR
\end{algorithmic}
\end{algorithm}



\section{Experiments}\label{sec:exp}
In this section, we conduct extensive evaluation of the proposed method on three widely used ReID datasets: Market1501~\cite{DBLP:conf/iccv/ZhengSTWWT15}, \mbox{DukeMTMC-ReID}~\cite{DBLP:conf/iccv/ZhengZY17} and CUHK03~\cite{DBLP:conf/cvpr/LiZXW14} in a \mbox{single-query} mode. DMIH is implemented with PyTorch~\cite{paszke2017automatic} on a NVIDIA M40 GPU server. We use the C++ implementation of MIH provided by the authors of~\cite{DBLP:journals/pami/0002PF14}\footnote{\url{https://github.com/norouzi/mih}} and conduct the retrieval experiments on a server with Intel Core CPU~(2.2GHz) and 96GB RAM.

\subsection{Datasets}

\paragraph{Market1501} Market1501 dataset consists of 32,688 bounding boxes of 1,501 persons from 6 cameras. These bounding boxes are cropped by the \mbox{deformable-part-model}~(DPM) detector~\cite{DBLP:journals/pami/FelzenszwalbGMR10}. 12,936 images of 751 persons are selected from the dataset as training set, and the remaining 750 persons are divided into test set with 3,368 query images and 19,732 gallery images.

\paragraph{DukeMTMC-ReID} DukeMTMC-ReID dataset is a subset of the \mbox{DukeMTMC} dataset~\cite{DBLP:conf/eccv/RistaniSZCT16} for \mbox{image-based} \mbox{re-identification}.  It consists of 36,411 images of 1,812 persons from 8 \mbox{high-resolution} cameras. The whole dataset  is divided into training set with 16,522 images of 702 persons and test set with 2,228 query images and 17,661 gallery images of the remaining 702 persons.

\paragraph{CUHK03} CUHK03 dataset contains 14,097 images of 1,467 persons from 6 surveillance cameras. This dataset provides both manually labeled pedestrian bounding boxes and bounding boxes detected by the DPM detector. For this dataset, we choose the labeled images for evaluation. To be more consistent with real application, we adopt the widely recognized~\cite{DBLP:conf/mm/WangYCLZ18, DBLP:conf/cvpr/ChangHX18, DBLP:conf/eccv/SunZYTW18} protocol proposed in~\cite{DBLP:conf/cvpr/ZhongZCL17}.

\subsection{Experimental Setup}

\paragraph{Baselines and Evaluation Protocol} Both hashing methods and real-valued methods are adopted as baselines for comparison. The \mbox{state-of-the-art} hashing methods for comparison include: { 1) \emph{\mbox{non-deep} hashing methods}: } COSDISH~\cite{DBLP:conf/aaai/KangLZ16}, SDH~\cite{DBLP:conf/cvpr/ShenSLS15}, KSH~\cite{DBLP:conf/cvpr/LiuWJJC12}, ITQ~\cite{DBLP:conf/cvpr/GongL11}, LSH~\cite{DBLP:conf/compgeom/DatarIIM04}; {2) \emph{deep hashing methods}: }PDH~\cite{DBLP:journals/tip/ZhuKZFT17}, \mbox{HashNet}~\cite{DBLP:conf/iccv/CaoLWY17}, DPSH~\cite{DBLP:conf/ijcai/LiWK16}. Among these baselines, PDH is designed specifically for ReID. DRSCH~\cite{DBLP:journals/tip/ZhangLZZZ15} is not adopted for comparison because it has been found to be outperformed by PDH. The \mbox{state-of-the-art} \mbox{real-valued} ReID methods for comparison include: { 1) \emph{metric learning methods}: }KISSME~\cite{DBLP:conf/cvpr/KostingerHWRB12}; { 2) \emph{deep learning methods}: }\mbox{pose-driven} deep convolutional~(PDC)~\cite{DBLP:conf/iccv/SuLZX0T17}, Spindle~\cite{DBLP:conf/cvpr/ZhaoTSSYYWT17}, MGN~\cite{DBLP:conf/mm/WangYCLZ18}.

Following the standard evaluation protocol on ReID tasks~\cite{DBLP:conf/mm/WangYCLZ18}, we report the mean average precision~(mAP) and Cumulated Matching Characteristic~(CMC) to verify the effectiveness of our proposed method. Furthermore, to verify the high accuracy and fast query speed DMIH can achieve, we adopt \mbox{precision-time} and \mbox{recall-time}~\cite{DBLP:journals/corr/Cai16b,DBLP:journals/corr/abs-1711-06016} to evaluate DMIH and baselines. Specifically, after constructing the MIH tables, we conduct \mbox{$k$-nearest} neighbor search by performing \mbox{multi-index} hash lookup with different $k$ and then do \mbox{re-ranking} on the selected nearest neighbors according to the corresponding deep features~(\mbox{2048-dim}) before the hash layer. Based on the re-ranking results, we choose the \mbox{top-20} nearest neighbors which have the minimum Euclidean distance to the query image and then calculate the precision and recall. At last, we summarize the time of \mbox{multi-index} hash lookup and \mbox{re-ranking} to draw the \mbox{precision-time} and \mbox{recall-time} curves~\cite{DBLP:journals/corr/Cai16b,DBLP:journals/corr/abs-1711-06016}.

\paragraph{Implementation Details} For DMIH, we set $\beta=2.0, \gamma=0.5$ and $T=160$ for all the experiments based on cross-validation strategy. We use Adam algorithm~\cite{DBLP:journals/corr/KingmaB14} for learning and choose the learning rate from $[10^{-5},10^{-3}]$. The initial learning rate is set to $4\times 10^{-4}$ and the weight decay parameter is set to $5\times 10^{-4}$. The \mbox{multi-branch} network based on \mbox{ResNet-50} is \mbox{pre-trained} on ImageNet dataset~\cite{DBLP:conf/cvpr/DengDSLL009}. The input for the image modality is raw pixels with the size of $384 \times 128$. We fix the size of each \mbox{mini-batch} to be 64, which is made up of 16 pedestrians and 4 images for each pedestrian.

For \mbox{non-deep} hashing methods, we use two image features. The first one is the Local Maximal Occurrence~(LOMO) feature~\cite{DBLP:conf/cvpr/LiaoHZL15}. After getting the LOMO feature, we use PCA to reduce the dimensionality to \mbox{3,000}. The second one is the deep features extracted by \mbox{ResNet-50} \mbox{pre-trained} on ImageNet. Among hashing baselines, KSH and SDH are \mbox{kernel-based} methods. For these methods, 1,000 data points are randomly selected from training set as anchors to construct kernels by following the suggestion of the original authors. For deep hashing methods, we adopt the same \mbox{multi-branch} network for a fair comparison. For all hashing based ReID methods, other hyper-parameters are set by following the suggestion of the corresponding authors. The source code is available for all baselines except PDH and MGN. We carefully re-implement PDH and MGN using PyTorch.

\subsection{Accuracy}
\paragraph{Comparison with Hashing Methods} We report the mAP on three datasets in Table~\ref{tab:mAP}, where ``COSDISH''/\mbox{``COSDISH+CNN''} denotes COSDISH with LOMO/deep features, respectively. Other notations are defined similarly. The CMC results are moved to supplementary materials due to space limitation. From Table~\ref{tab:mAP}, we can see that DMIH outperforms all baselines including deep hashing based ReID methods, deep hashing methods and \mbox{non-deep} hashing methods in all cases.
\begin{table*}[tb]
\centering
\scriptsize
\caption{mAP~(in percent). The best accuracy is shown in boldface.}
\scalebox{0.87}{
\label{tab:mAP}
\begin{tabular}{c|c|c|c|c|c|c|c|c|c|c|c|c}
\Xcline{1-13}{1pt}
\multirow{2}{*}{Method} & \multicolumn{4}{c|}{{Market1501}} & \multicolumn{4}{c|}{{DukeMTMC-ReID}}& \multicolumn{4}{c}{{CUHK03}}\\
\cline{2-13} &  32 bits & 64 bits & 96 bits & 128 bits &  32 bits & 64 bits & 96 bits & 128 bits &  32 bits & 64 bits & 96 bits & 128 bits\\
\hline
DMIH           & {\bf 31.41} & {\bf 49.80} & {\bf 58.14} & {\bf 62.24} & {\bf 24.29} & {\bf 40.88} & {\bf 47.70} & {\bf 52.35} & {\bf 24.95} & {\bf 40.28} & {\bf 44.82} & {\bf 48.75} \\
\hline
PDH            & {24.75} & {38.73} & {46.07} & {50.76} & {16.30} & {30.79} & {39.16} & {43.48} & {18.72} & {31.27} & {35.10} & {41.02} \\
HashNet        & {13.10} & {22.23} & {25.53} & {26.26}  & {8.10} & {13.53} & {15.74} & {18.41} & {12.79} & {16.39} & {17.83} & {18.27} \\
DPSH           & {12.34} & {20.27} & {24.68} & {29.42} & {12.87} & {20.02} & {26.45} & {29.42} & {11.22} & {16.11} & {19.72} & {20.82} \\
\hline
COSDISH   &  {1.89} & {3.68} & {4.83} & {5.94} & {1.02} & {2.39} & {3.81} & {5.11} & {0.82} & {1.54} & {2.59} & {3.01}\\
SDH       &  {1.65} & {2.93} & {3.78} & {4.06} & {0.98} & {1.89} & {2.25} & {2.42} & {1.00} & {1.24} & {1.32} & {1.65}\\
KSH       &  {4.66} & {5.62} & {6.16} & {6.20} & {2.13} & {2.67} & {3.31} & {3.34} & {2.86} & {2.53} & {2.11} & {1.75}\\
ITQ       &  {1.70} & {3.00} & {3.83} & {4.43} & {0.91} & {1.41} & {1.77} & {2.16} & {0.68} & {0.76} & {0.82} & {0.95}\\
LSH       &  {0.44} & {0.83} & {1.18} & {1.68} & {0.40} & {0.58} & {0.83} & {1.06} & {0.37} & {0.46} & {0.44} & {0.68}\\
\hline
COSDISH+CNN    &  {0.79} & {1.06} & {1.47} & {1.82} & {0.62} & {1.09} & {1.42} & {1.79} & {0.39} & {0.57} & {0.63} & {0.62}\\
SDH+CNN        &  {0.73} & {1.26} & {1.55} & {1.67} & {0.66} & {0.89} & {1.06} & {1.40} & {0.44} & {0.65} & {0.63} & {0.63}\\
KSH+CNN        &  {0.77} & {0.74} & {0.54} & {0.68} & {0.30} & {0.37} & {0.44} & {0.46} & {0.49} & {0.41} & {0.33} & {0.41}\\
ITQ+CNN        &  {0.77} & {1.07} & {1.21} & {1.29} & {0.56} & {0.92} & {1.21} & {1.34} & {0.38} & {0.38} & {0.43} & {0.45}\\
LSH+CNN        &  {0.50} & {0.77} & {1.04} & {1.27} & {0.48} & {0.74} & {0.99} & {1.17} & {0.33} & {0.35} & {0.35} & {0.42}\\
\Xcline{1-13}{1pt}
\end{tabular}}
\end{table*}

\begin{figure*}[tb]
\centering
\includegraphics[scale=0.5]{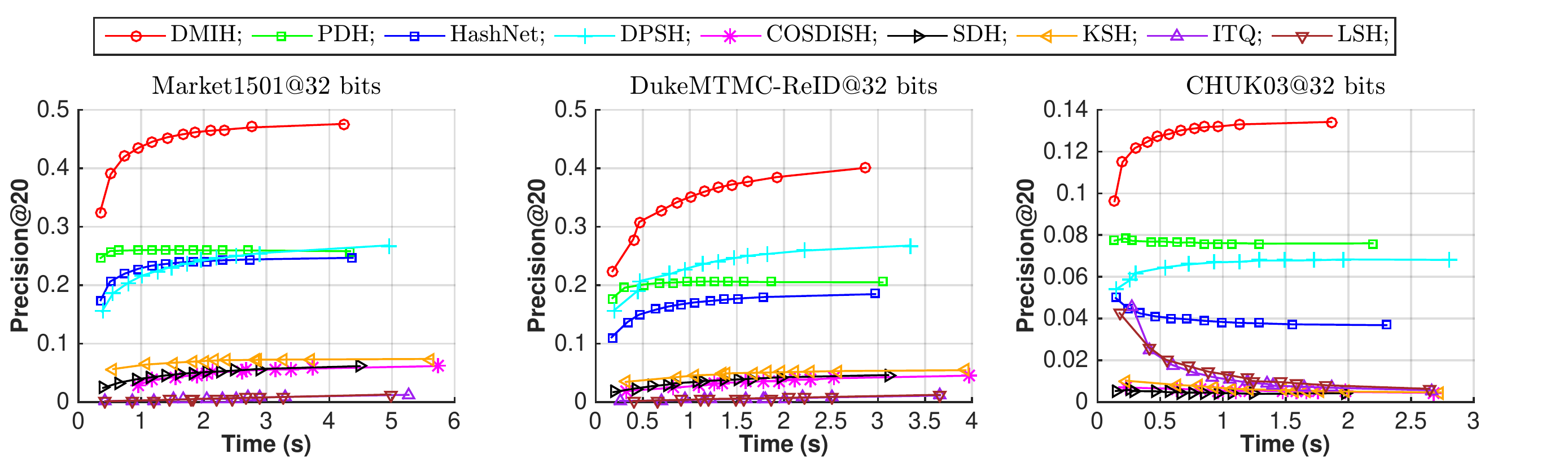}\vspace{-10pt}
\caption{Precision-time curves on three datasets with binary code length being 32 bits.}\vspace{-15pt}
\label{fig:precision20-32}
\end{figure*}

Furthermore, we also present the \mbox{precision-time} curves on three datasets in Figure~\ref{fig:precision20-32}. Due to the mAP results in Table~\ref{tab:mAP}, \mbox{non-deep} hashing methods utilize LOMO features. The \mbox{recall-time} curves and more \mbox{precision-time} curves with other bits are moved to the supplementary materials due to space limitation. From Figure~\ref{fig:precision20-32}, we can find that DMIH can achieve the highest precision among all the hash methods while costs less time in all cases. Hence, \mbox{DMIH} can significantly outperform existing \mbox{non-deep} hashing methods and deep hashing methods in terms of both \emph{efficiency} and \emph{accuracy}. In addition, we find that the methods with higher mAP or CMC do not necessarily have better \mbox{precision-time} and \mbox{recall-time}. That is to say, only using mAP and CMC to evaluate hashing methods may not be comprehensive. So we adopt mAP, CMC, \mbox{precision-time} and \mbox{recall-time} to comprehensively verify the promising efficiency and accuracy of our method.
\paragraph{Comparison with Real-Valued ReID Methods} We also compare our DMIH with some representative \mbox{real-valued} ReID methods. As the dimension of the \mbox{real-valued} features is usually high, we increase the binary code length of DMIH for fair comparison. We report CMC@20 and the corresponding retrieval time in Table~\ref{tab:compare_with_ReID_methods}, where the results of BoW+KISSME~\cite{DBLP:conf/cvpr/KostingerHWRB12}, Spindle~\cite{DBLP:conf/cvpr/ZhaoTSSYYWT17} and PDC~\cite{DBLP:conf/iccv/SuLZX0T17} are directly copied from the original papers and ``--" denotes that the result of the corresponding setting is not reported in the original papers. \mbox{``DMIH(RK)~(32 bits)''} denotes the DMIH method of 32 bits with \mbox{re-ranking} after hash lookup. Other variants of DMIH are named similarly. The retrieval time includes the time for both hash lookup and \mbox{re-ranking}. Furthermore, we also report the speedup of DMIH relative to the best baseline MGN.

From Table~\ref{tab:compare_with_ReID_methods}, we can see that DMIH can still achieve the best performance in all cases when compared with \mbox{state-of-the-art} \mbox{real-valued} ReID methods. In particular, \mbox{DMIH} can outperform the best real-valued baseline MGN in terms of both accuracy and efficiency with suitable code length. We can also find that DMIH can achieve higher accuracy by increasing binary code length. However, longer binary code typically leads to worse retrieval efficiency. In real applications, one can choose proper binary code length to get a good \mbox{trade-off} between efficiency and accuracy.
\begin{table*}[tb]
\centering
\scriptsize
\caption{CMC@20~(in percent) and Time~(in second). The first/second best result is shown in boldface/underline.}
\label{tab:compare_with_ReID_methods}
\scalebox{0.95}{
\begin{tabular}{l|c|c|c|c|c|c|c|c|c}
\Xcline{1-10}{1pt}
\multirow{2}{*}{Method} & \multicolumn{3}{c|}{{Market1501}} & \multicolumn{3}{c|}{{DukeMTMC-ReID}}& \multicolumn{3}{c}{{CUHK03}}\\
 \cline{2-10} & CMC@20 & Time & Speedup & CMC@20 & Time & Speedup & CMC@20 & Time & Speedup\\
\hline
DMIH(RK)~(32 bits)    & 97.4  & 2.34s & 48$\times$ & 91.8 & 1.92s & 43$\times$ & 85.6 & 0.48s & 32$\times$\\
DMIH(RK)~(96 bits)    & 98.3  & 4.16s & 27$\times$ & 94.3 & 3.44s & 24$\times$ & 92.1 & 0.93s & 17$\times$\\
DMIH(RK)~(256 bits)   & 98.4  & 7.66s & 15$\times$ & 94.8 & 5.39s & 15$\times$ & 92.9 & 1.54s & 10$\times$\\
DMIH(RK)~(512 bits)   & 98.6  & 11.26s & 10$\times$ & 95.4 & 9.07s & 9$\times$ & 93.2 & 2.17s & 7$\times$\\
DMIH(RK)~(1024 bits)  & {\bf 98.8} & 14.34s & 8$\times$ & 95.8 & 11.00s & 8$\times$ & \underline{93.4} & 2.75s & 6$\times$\\
DMIH(RK)~(2048 bits)  & \underline{98.7} & 20.12s & 6$\times$ & {\bf 96.3} & 15.86s & 5$\times$ & {\bf 93.5} & 4.19s & 4$\times$\\
 \hline
BoW+KISSME~\cite{DBLP:conf/cvpr/KostingerHWRB12}        & 78.5 & {--} & -- & {--} & -- & {--} & -- &-- &-- \\
\hline
Spindle~\cite{DBLP:conf/cvpr/ZhaoTSSYYWT17}           & 96.7 & {--} & -- & {--} & -- & {--} & -- &-- &-- \\
PDC~\cite{DBLP:conf/iccv/SuLZX0T17} & 96.8 & {--} & -- & {--} & -- & {--} & -- &-- &-- \\
MGN~\cite{DBLP:conf/mm/WangYCLZ18} & \underline{98.7} & {113.26s} & 1$\times$ & \underline{95.9} & {82.79s} & 1$\times$ & 92.2 & {15.73s} & 1$\times$\\
\Xcline{1-10}{1pt}
\end{tabular}}
\end{table*}

\subsection{Ablation Study}
\begin{figure}[tb]
  \begin{minipage}[b]{0.5\textwidth}
    \centering
    \begin{tabular}{c@{ }@{ }c@{ }@{ }c@{ }@{ }c}
	\begin{minipage}{0.48\linewidth}\centering
    \includegraphics[width=1\textwidth]{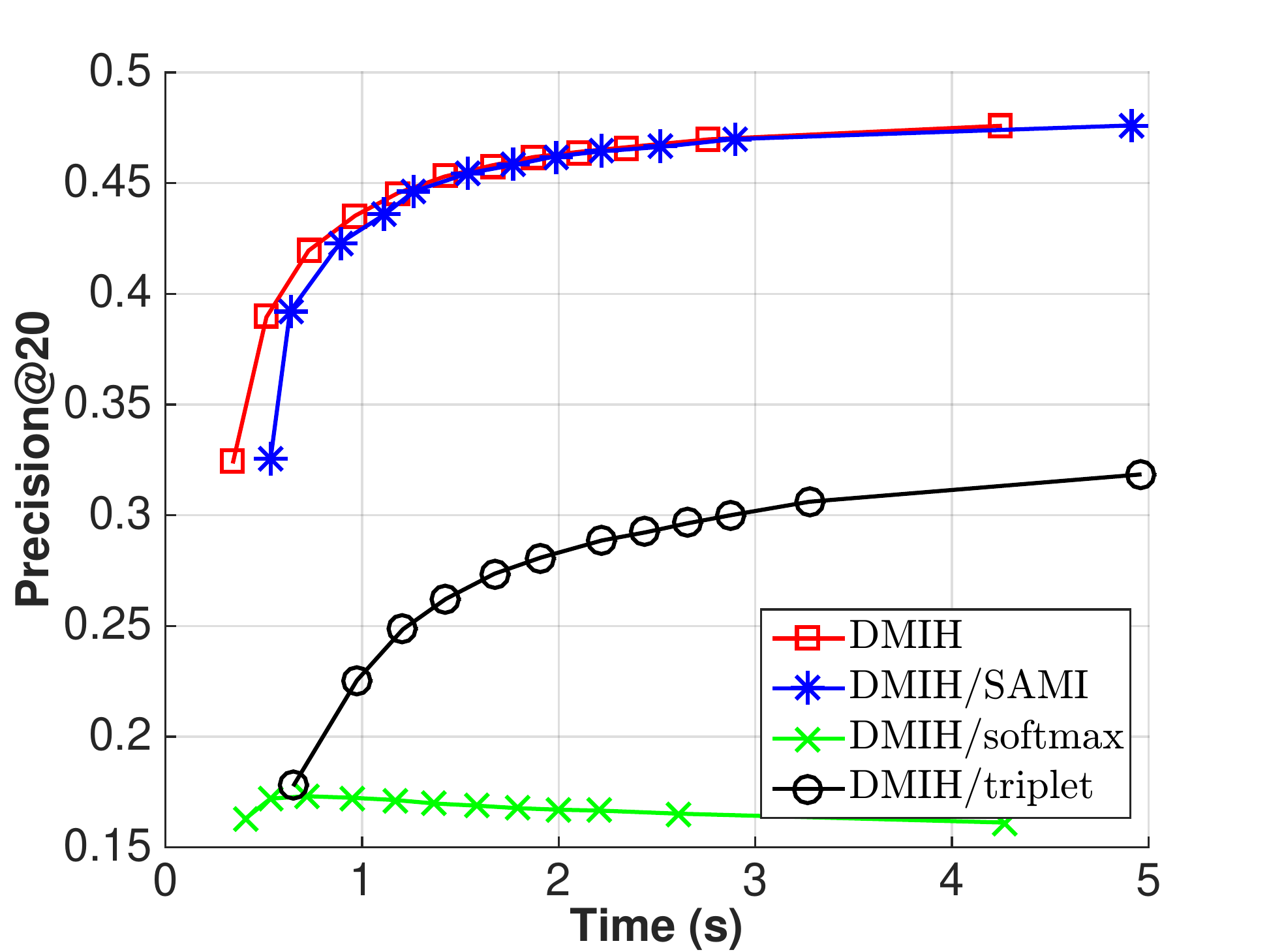}\\
    {(a) Market1501@32 bits}
	\end{minipage} &
	\begin{minipage}{0.48\linewidth}\centering
    \includegraphics[width=1\textwidth]{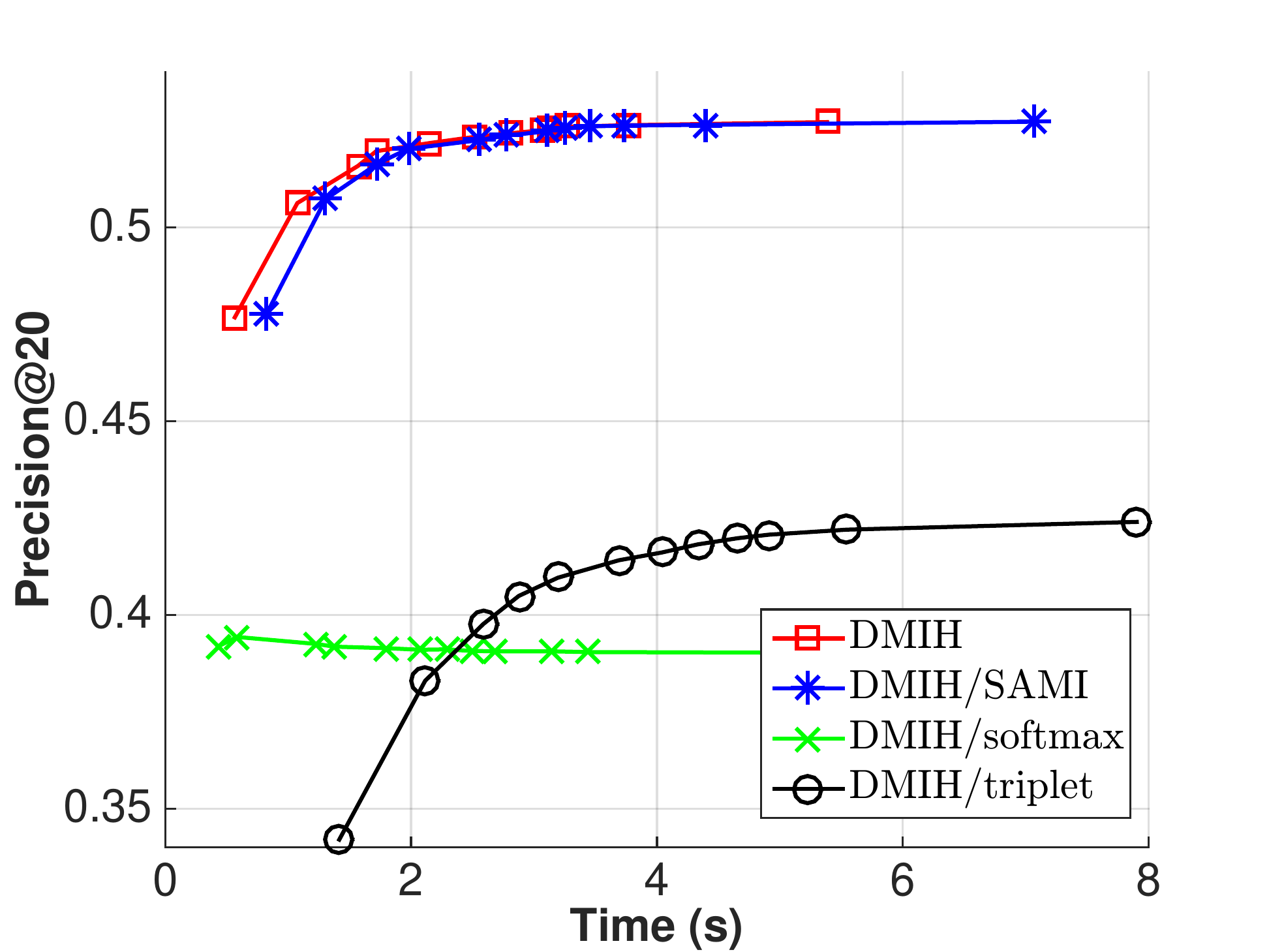}\\
    {(b) Market1501@96 bits}
	\end{minipage}
	\end{tabular}
	\caption{Ablation study. }
	\label{fig:hyper-parameter}
  \end{minipage}%
  \begin{minipage}[b]{0.5\textwidth}
    \centering
    \begin{tabular}{c@{ }@{ }c@{ }@{ }c@{ }@{ }c}
	\begin{minipage}{0.48\linewidth}\centering
    \includegraphics[width=1\textwidth]{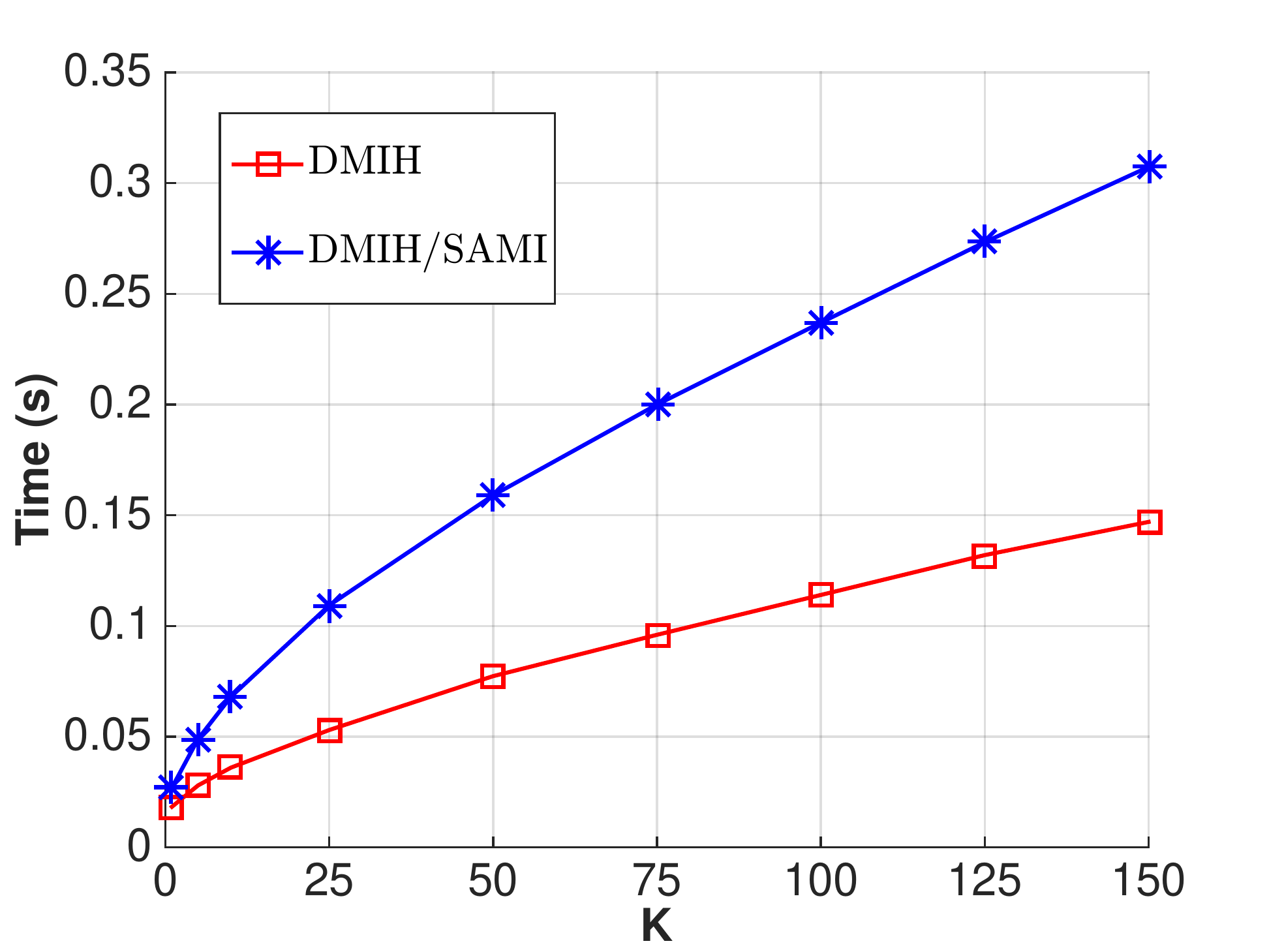}\\
    {(a) Market1501@32 bits}
	\end{minipage} &
	\begin{minipage}{0.48\linewidth}\centering
    \includegraphics[width=1\textwidth]{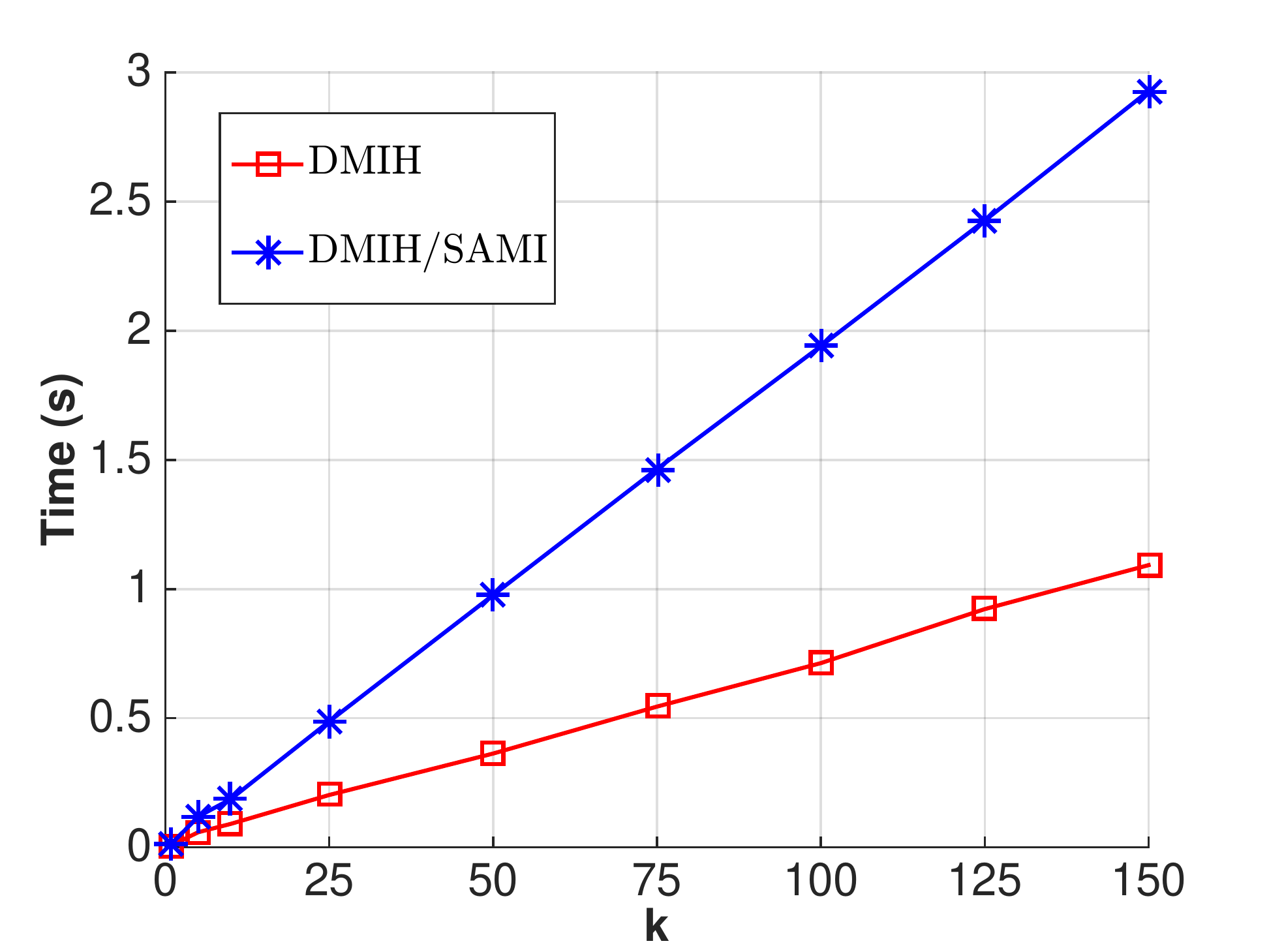}\\
    {(b) Market1501@96 bits}
	\end{minipage}
	\end{tabular}
    \caption{Time cost obtained by $k$NN.}
    \label{fig:speedup}
  \end{minipage}\vspace{-10pt}
\end{figure}

We conduct experiments to study whether all the loss terms in DMIH are necessary by removing these loss terms separately. The result on Market1501 dataset with 32 bits and 96 bits is presented in Figure~\ref{fig:hyper-parameter}. Here, ``DMIH/SAMI'' denotes the DMIH variant without the SAMI loss $\LM_s(\cdot)$, i.e., $\gamma=0$, and other notations are defined similarly. We can find that softmax loss and triplet loss can significantly improve accuracy. Furthermore, by comparing DMIH with DMIH/SAMI, we can find that SAMI loss can further accelerate the query speed without losing accuracy. As the time shown in Figure~\ref{fig:hyper-parameter} contains the hash table lookup time and re-ranking time, we compare the hash lookup time separately of DMIH with DMIH/SAMI in Figure~\ref{fig:speedup}. From Figure~\ref{fig:speedup}, we can see that DMIH can achieve $2\sim3$ times acceleration of hash lookup with the help of SAMI loss.

\subsection{Effect of Block-Wise MIH Table Construction}
\begin{wrapfigure}{r}{0.5\textwidth}\vspace{-18pt}
\centering
\small
\begin{tabular}{c@{ }@{ }c@{ }@{ }c@{ }@{ }c}
\begin{minipage}{0.48\linewidth}\centering
    \includegraphics[width=1\textwidth]{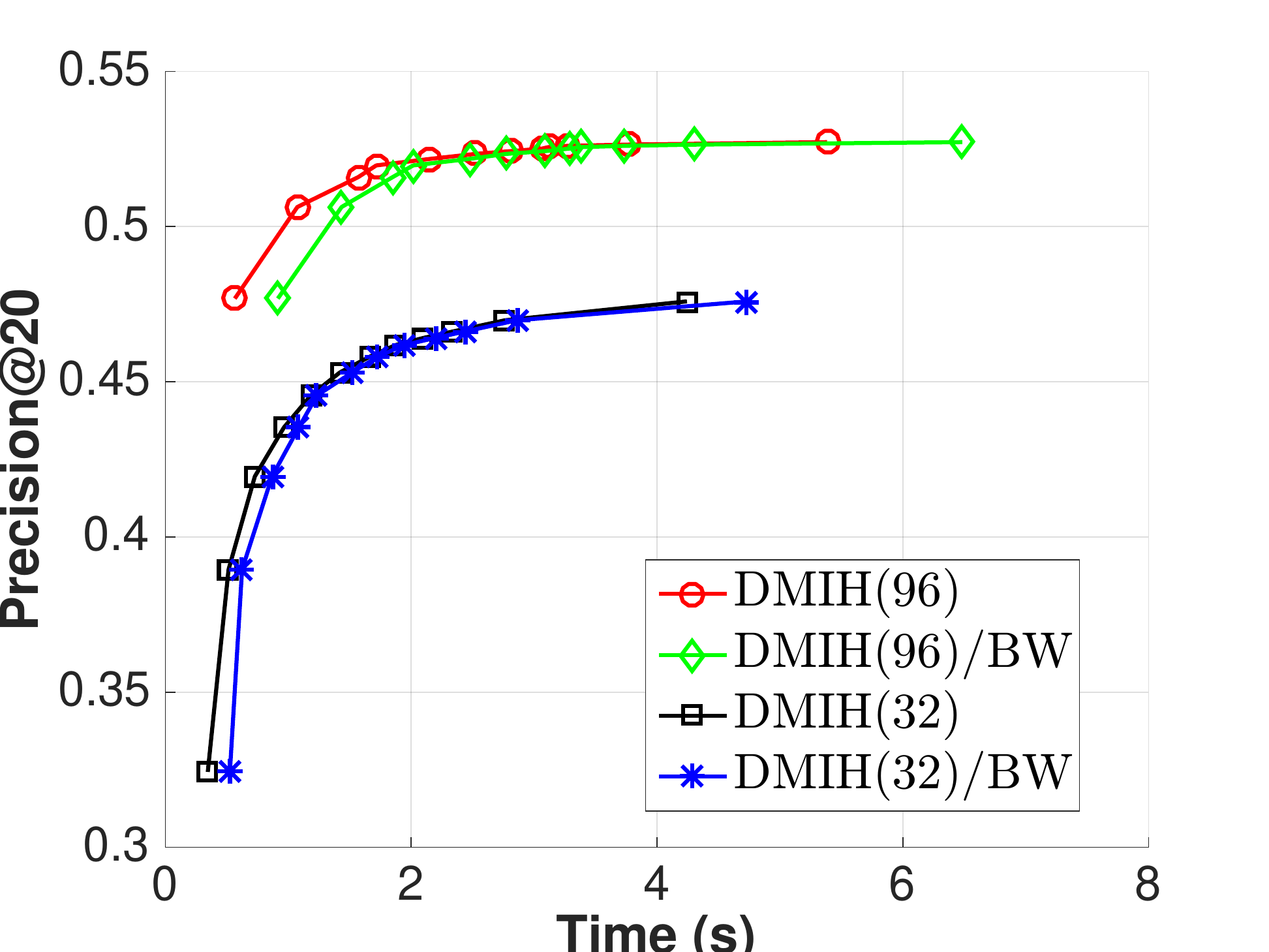}\\
    {(a) Market1501@32, 96 bits}
\end{minipage} &
\begin{minipage}{0.48\linewidth}\centering
    \includegraphics[width=1\textwidth]{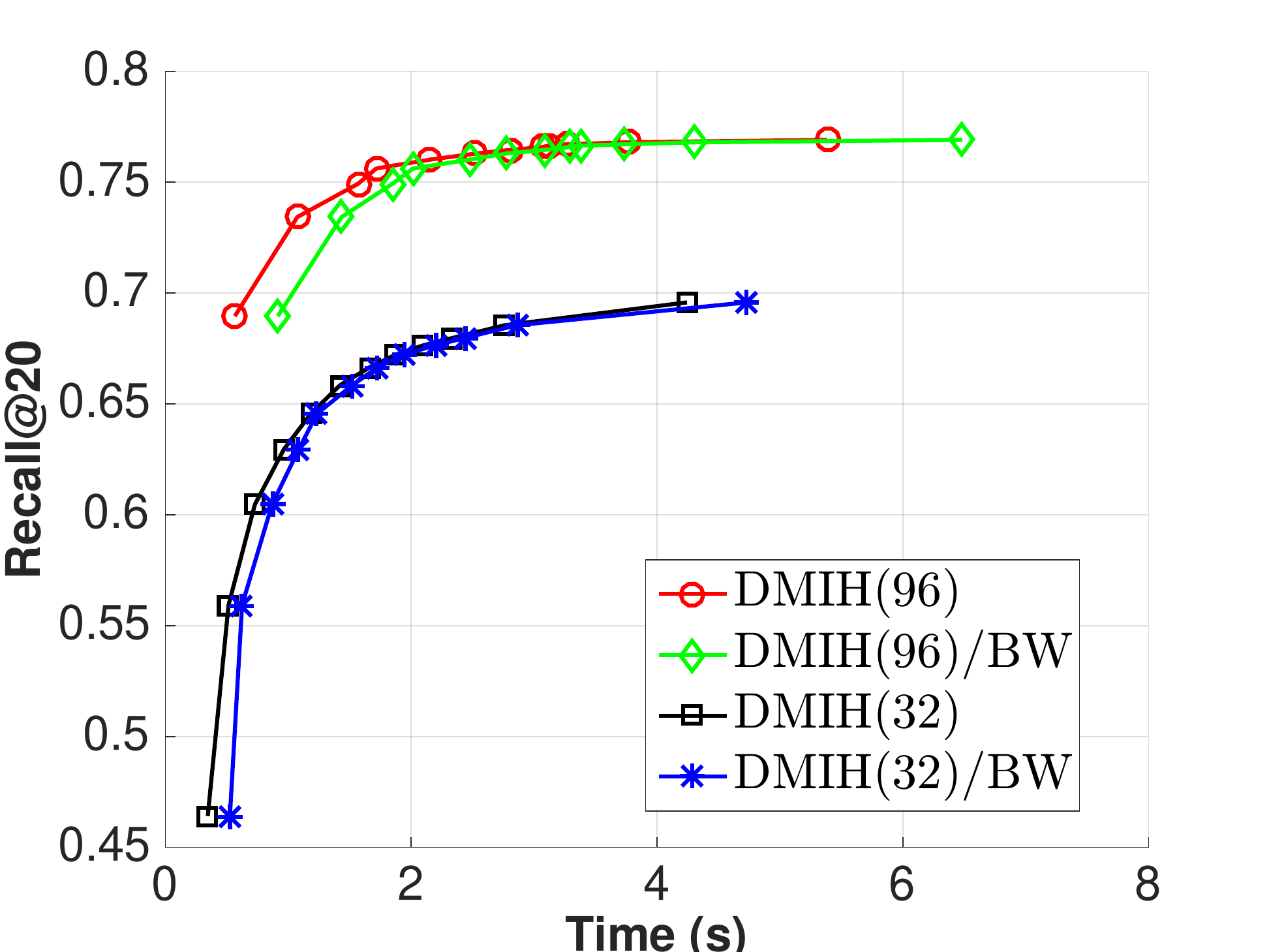}\\
    {(b) Market1501@32, 96 bits}
\end{minipage}
\end{tabular}
\caption{Effect of block-wise MIH table construction strategy.}\vspace{-10pt}
\label{fig:blockwise}
\end{wrapfigure}
To verify the effectiveness of the proposed \mbox{block-wise} MIH tables construction strategy, we compare DMIH with a variant without the \mbox{block-wise} MIH table construction strategy. More specifically, for the variant without the \mbox{block-wise} MIH table construction strategy, we directly divide the learned binary code $[\d_i;\g_i;\h_i]^T$ into $m$ \mbox{sub-binary} codes to construct MIH tables. For fair comparison, we utilize the same learned binary code for both DMIH and the variant. The \mbox{precision-time} and \mbox{recall-time} curves on Market1501 dataset with binary code length being 32 bits and 96 bits are shown in Figure~\ref{fig:blockwise}, where ``DMIH(32)/BW'' denotes the variant without the \mbox{block-wise} MIH construction strategy and other notations are defined similarly. From Figure~\ref{fig:blockwise}, we can see that using block-wise strategy can get a speedup in retrieval efficiency without losing accuracy.

%

\section{Conclusion}\label{sec:conclusion}
In this paper, we propose a novel deep hashing method, called DMIH, for person ReID. DMIH is an \mbox{end-to-end} deep learning framework, which integrates \mbox{multi-index} hashing and \mbox{multi-branch} based networks into the same framework. Furthermore, we propose a novel \mbox{block-wise} \mbox{multi-index} hashing tables construction strategy and a \mbox{search-aware} \mbox{multi-index} loss to further improve the search efficiency. Experiments on real datasets show that DMIH can outperform other baselines to achieve the \mbox{state-of-the-art} retrieval performance in terms of both {efficiency} and {accuracy}.

\small
\bibliography{ref}
\bibliographystyle{abbrv}

\end{document}